\documentclass[conference]{IEEEtran}
\IEEEoverridecommandlockouts

\usepackage{cite}
\usepackage{amsmath,amssymb,amsfonts}
\usepackage{algorithmic}
\usepackage{graphicx}
\usepackage{textcomp}
\usepackage{xcolor}

\usepackage{overpic}
\usepackage[pagebackref=true,breaklinks=true,colorlinks,bookmarks=false]{hyperref}

\graphicspath{{figures/}}

\begin{document}

\newcommand\figref{Figure~\ref}

\newcommand\tabref{Table~\ref}

\title{
Localization matters too: How localization error affects UAV flight\\

}

\author{\IEEEauthorblockN{Suquan Zhang$^{1*}$, Yuanfan Xu$^{1*}$, Shu'ang Yu$^{1}$, Qingmin Liao$^{2}$, Jincheng Yu$^{1\dagger}$ and Yu Wang$^{1\dagger}$}

\thanks{$^{1}$ Department of Electronic Engineering, Tsinghua
University, Beijing, China. \{zhangsq23, xuyf20, yusa20\}@mails.tsinghua.edu.cn \{yu-jc, yu-wang\}@tsinghua.edu.cn }
\thanks{$^{2}$ Shenzhen International Graduate School, Tsinghua University, Beijing, China. liaoqm@tsinghua.edu.cn}
\thanks{* These authors contributed equally.}
\thanks{$^\dagger$ Corresponding Authors.}
}

\maketitle

\begin{abstract}
The maximum safe flight speed of a Unmanned Aerial Vehicle (UAV) is an important indicator for measuring its efficiency in completing various tasks. 
This indicator is influenced by numerous parameters such as UAV localization error, perception range, and system latency. However, in terms of localization errors, although there have been many studies dedicated to improving the localization capability of UAVs, there is a lack of quantitative research on their impact on speed.
In this work, we model the relationship between various parameters of the UAV and its maximum flight speed. 
We consider a scenario similar to navigating through dense forests, where the UAV needs to quickly avoid obstacles directly ahead and swiftly reorient after avoidance. 
Based on this scenario, we studied how parameters such as localization error affect the maximum safe speed during UAV flight, as well as the coupling relationships between these parameters.
Furthermore, we validated our model in a simulation environment, and the results showed that the predicted maximum safe speed had an error of less than $20\%$ compared to the test speed. In high-density situations, localization error has a significant impact on the UAV's maximum safe flight speed. This model can help designers utilize more suitable software and hardware to construct a UAV system.
\end{abstract}


\section{Introduction}

UAVs have always been a hot topic in the field of robotics, and with the advancement of technology [1]-[4], an increasing variety of UAVs are being extensively utilized in various scenarios. 

These applications often take place in challenging environments or demand efficient task completion. The importance of UAV maneuverability cannot be overstated in such scenarios [5]. 

Therefore, the maximum safe flying speed of UAVs becomes a crucial indicator. It ensures that UAVs can swiftly and safely carry out their missions while maintaining control and avoiding potential hazards.

It is evident that UAVs with different hardware or algorithms may demonstrate varying levels of efficiency in different tasks [6]. In order to enhance mission execution efficiency, UAVs need to be capable of flying at high speeds while simultaneously conducting rapid obstacle detection and avoidance in the environment. During this process, factors such as algorithm latency,  sensing range, and localization error can significantly impact the maximum safe speed of UAVs [5].

Currently, we have an increasing array of sensors and algorithms to choose from [7]-[9], which collectively form the entire system of a UAV. However, it is evident that few modules excel in all aspects. For example, increasing the sensing range often leads to an increase in latency. As a result, it becomes necessary to make trade-offs during the module selection process.

Although it is widely acknowledged that the localization error, sensing range, and overall system latency of UAVs can impact their speed, few people can quantitatively describe the extent of these parameters' influence.

In particular, regarding localization error, although there has been a lot of related work in improving the localization capability of UAVs [10]-[12], determining how they quantitatively affect the maximum flight speed of UAVs remains a question. This leads to difficulties in evaluating the final performance when selecting new sensors or algorithms, especially when two or more variables change simultaneously (for example, reducing localization error accompanied by an increase in localization latency).

In this article, we aim to uncover the influence of various parameters, particularly localization error, on the maximum flight speed of UAVs. We approach this by modeling the relationship between the UAV's parameters and the maximum flight speed from a theoretical analysis perspective. Additionally, we analyze the coupling relationships among these parameters. 

\begin{figure*}[htbp]
   \begin{overpic}[width=\textwidth]{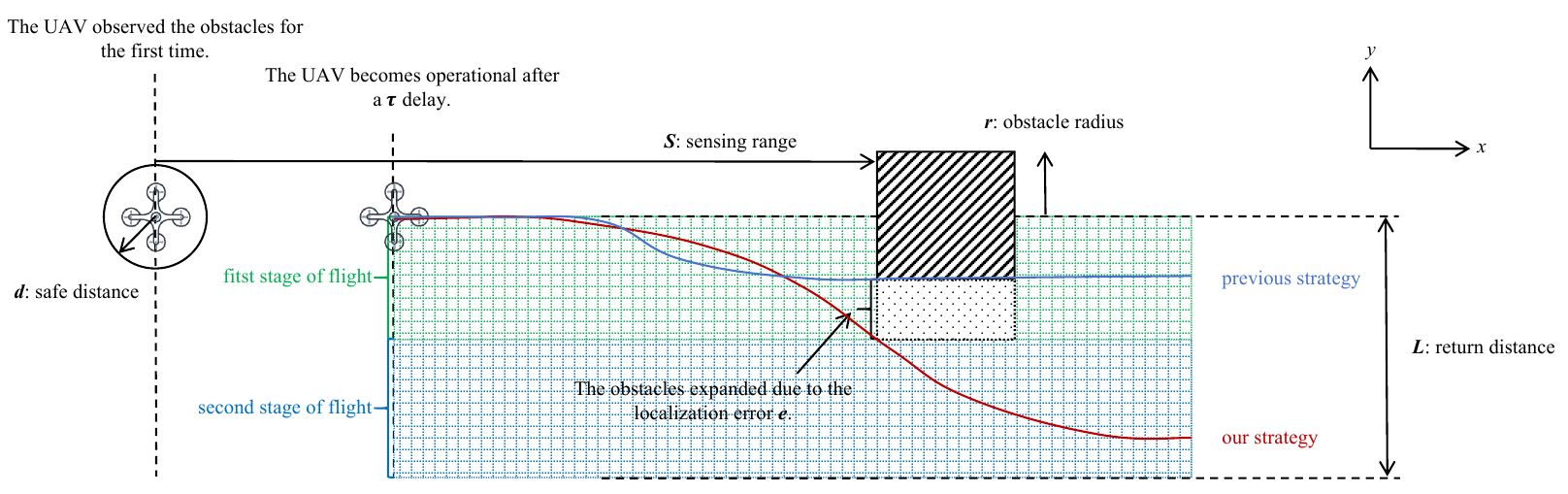} \small
   \end{overpic}
   \caption{Compared to previous strategy [13], our strategy requires the UAV to avoid obstacles directly ahead in the most extreme manner possible, and to realign its direction within a limited distance $L$. Additionally, we also take into consideration the expansion of obstacles in the UAV's map caused by localization error.
   }\label{fig:scene}
\end{figure*}

\subsection{Related Works}

The improvement of the maximum safety speed of UAVs has always been a key indicator for various algorithms [14], [15]. As an important component of UAV systems, the enhancement of localization accuracy holds significant importance for enabling UAVs to achieve faster flight and control. In [16], [17], researchers have designed high-precision and high-frequency localization systems to realize high-speed flight of UAVs in complex scenarios.

In the field of UAV localization methods, many previous studies have also investigated different approaches for UAV localization. These methods can assist UAVs in achieving robust and rapid localization using various sensors. For example, in [10], the authors proposed a low-cost robust localization algorithm based on a monocular vision system. In [12], point features and line features were utilized to achieve high-precision and low-latency localization on a public dataset. Furthermore, some works have also focused on accelerating localization systems from a hardware perspective [18].

However, in these works, the effectiveness of localization systems is often evaluated based on the accuracy and latency of the localization itself, making it difficult to directly relate them to the maximum safety speed of UAVs. Researchers struggle to intuitively understand how reducing localization error can benefit UAVs in achieving higher speeds. There is still a lack of a quantified model that establishes the relationship between localizaiton errors and UAV speed to aid in UAV design.

There are also some studies that explore the theoretical relationship between the maximum speed of UAVs and various parameters. The earliest modeling of UAVs in this context can be traced back to [19]. The authors constrained the UAV's flight based on collision considerations, assuming that the UAV must come to a stop within a fixed distance with maximum acceleration.

Under this assumption, the stopping distance of the UAV can be calculated at maximum acceleration and serves as a useful indicator of the UAV's flight speed in collision-free scenarios. However, in practical flight operations, it is not desirable for the UAV to abruptly stop in front of obstacles. A smoother trajectory is often a fundamental criterion in UAV path planning.

The work presented in [13] addresses this issue and provides a more detailed explanation of the impact of sensing range and latency. With a focus on achieving smooth UAV flight, this study suggests that the UAV should maintain a constant speed in the target flight direction and utilize vertical acceleration alone to navigate around obstacles.  
It assumes that the UAV applies vertical acceleration the moment it detects the obstacle and provides reverse acceleration when halfway through the process. This allows the UAV to fly closely along the obstacle, as illustrated in \figref{fig:scene}. While this model considers the size of the obstacle and the smoothness of flight, it overlooks the influence of obstacle density on flight speed. Additionally, the paper does not provide an explanation for the impact of localization error on flight.

In summary, previous studies have not adequately explained the impact of localization error on the maximum safe speed of UAVs, nor have they fully discussed the coupling relationships among various parameters.

\subsection{Contribution}

In this work, we discuss the impact of parameters such as localization error, sensing range, algorithm delay, and physical limitations of UAVs on the maximum flight speed of UAVs in UAV obstacle avoidance scenarios. The basic scene setting is shown in \figref{fig:scene}.

We illustrate the relationship between the maximum safe speed and these variables using mathematical expressions. Additionally, we provide insights into the trends of the maximum safe speed under different parameter values and discuss the sensitivity of each parameter under varying conditions. By comparing different existing UAV hardware modules, we demonstrate how this tool can be utilized for UAV design analysis. The results show that, through a thoughtful combination of components and an analyzed hardware configuration, higher UAV safe flight speeds can be achieved compared to a brute-force approach of simply stacking performance components.

We conducted tests on this model in a simulation environment, and the results indicate that even at a maximum flight speed of $18m/s$, our model can still ensure that the predicted flight speed error is within $20\%$.

This work is the first to comprehensively address the influence of UAV localization error, sensing range, and algorithm latency on the maximum safe speed of UAVs from a theoretical modeling perspective.

\section{Method}

\subsection{Important assumptions}

In this article, we make certain assumptions regarding the decision-making and localization aspects of the UAV.

Firstly, the UAV maintains a constant speed in the x-direction while flying at a fixed height. To avoid obstacles, it generates a force in the y-direction.

Regarding the UAV's planning and decision-making, it operates in real-time and relies only on current observations. The goal is to generate smoother curved flights by using the minimum necessary acceleration to avoid obstacles. The UAV aims to fly as close as possible to obstacles while departing. This stage is referred to as the first stage of flight.

Compared to previous works, we believe that in the second stage of flight, when the UAV has cleared the obstacle, it should quickly make a directional adjustment to return to its original heading within a sufficiently short distance $L$, \figref{fig:scene} illustrates the differences between two strategies.

\begin{figure}[htbp]
\centerline{\includegraphics[width=0.5\textwidth]{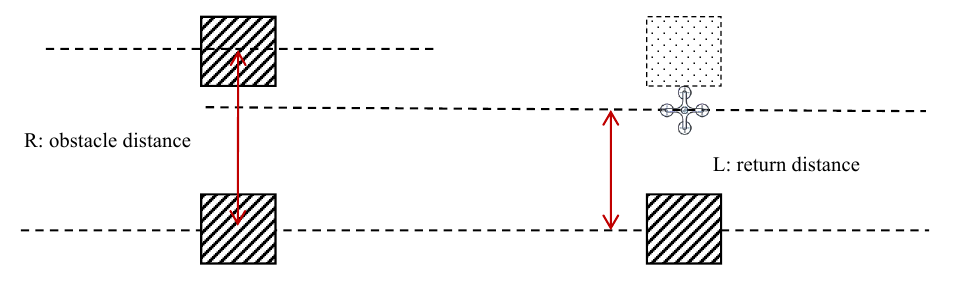}}
\caption{The distance $L$ for the UAV to realign its direction can be considered to have a simple mathematical relationship with the distance $R$ to the obstacle.}
\label{fig:distance}
\end{figure}

Here, we assume that the return distance $L$ has a simple mathematical relationship with the obstacle density as commonly understood. As shown in \figref{fig:distance}, if we denote the average distance between obstacles as $R$, we can obtain the following relationship:

\begin{equation}
L = R - r - d 
\end{equation}

Secondly, in terms of localization, we assume the UAV maintains mapping capabilities. Therefore, localization error can affect the perceived volume of obstacles in the map, resulting in the gradual expansion of obstacles. 

For micro-sized UAVs, the weight limitation prevents the installation of a LiDAR sensor. As a result, visual-inertial odometry (VIO) based on the fusion of visual and inertial sensors is a commonly used localization method [10], [12].

In this system, the estimation of the pose accumulates gradually, and the magnitude of the error increases continuously with the length of the trajectory [20].

In this study, We model the accumulated error as a deviation value $\Delta y = e\Delta x$, where $e$ is used to measure the magnitude of the error, this means that as the UAV moves in the x-direction, the observed deviation of obstacles in the y-direction gradually increases.

To simplify the influence of the UAV's geometric structure on its flight, we discuss the UAV's flight in configuration space [21], treating the UAV as a point mass. In this context, we can consider that the obstacles expand outward by a distance s, which represents the UAV's safety distance or geometric size. Theoretically, a rectangular obstacle, when expanded, would result in a rounded rectangle. However, for simplicity, we neglect the minor differences at the corners and still model it as a regular rectangle.

\begin{figure}[htbp]
\centerline{\includegraphics[width=0.5\textwidth]{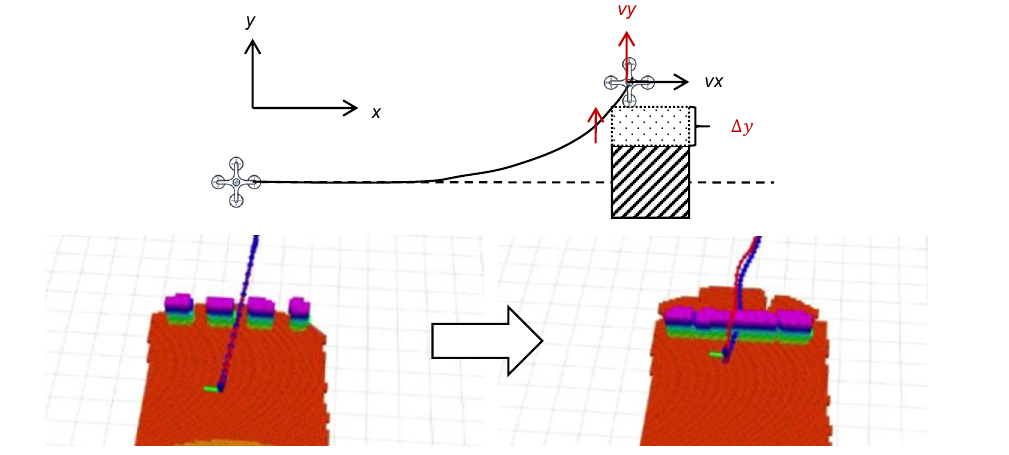}}
\caption{The localization error of the UAV causes the obstacles to expand in configuration space [21].}
\label{fig:error}
\end{figure}

Taking into account the influence of localization error, as the UAV flies towards obstacles, its error also accumulate. Since Visual-Inertial Odometry (VIO) is used for localization, the error accumulation starts from the moment the UAV perceives the obstacles. During the flight, the obstacles in the UAV's constructed map expand in the y-direction, gradually transforming into elongated rectangles as shown in \figref{fig:error}. The rate of expansion is related to the design of the error, as indicated by the previous relation $\Delta y = e\Delta x$. The expansion speed of the obstacles can be represented as $ev_x$, where $e$ represents the localization error.

For the computational latency of the UAV, we approximate the overall latency as the sum of delays from different components, with the main contributions coming from sensing and localization, as shown in Equation \eqref{eq:latency}.

\begin{equation} \label{eq:latency}
\tau = \tau_{depth} + \tau_{position} + ... 
\end{equation}

Finally, in terms of physical performance, we make the simplifying assumption that the UAV can fly with a given acceleration within a certain range($a<a_{max}$), and that the velocity and acceleration of the UAV exhibit a linear relationship.

For the change in acceleration, we consider that the rate of change of acceleration should also not exceed the physical limitations of the UAV, i.e., $jerk < j_{max}$. This mainly ensures that the transition of the UAV from maximum positive acceleration to maximum negative acceleration does not happen too quickly, preventing control failure.

\subsection{Symbol Explanation}

Due to the large number of parameters involved in this paper, we list all symbols and their corresponding meanings in \figref{fig:scene} and \tabref{tab1}, which represent the possible symbols that appear in the solution process.

\begin{table}[htbp]
\caption{Meanings of Symbols in the Paper}
\begin{center}

\begin{tabular}{|c|c|}
\hline
\textbf{ Symbol } & \textbf{meaning}  \\
\hline 
\textbf{$v_x$} & Velocity in the x direction of the UAV\\
\cline{1-2}
\textbf{$a_{max},j_{max}$} & The maximum acceleration and jerk of the UAV\\
\cline{1-2}
\textbf{$T$} & Flight time of the UAV in the first stage \\
\cline{1-2}
\textbf{$T'$} & $T-\tau$ \\
\cline{1-2}
\textbf{$t$} & The current moment \\
\cline{1-2}
\textbf{$t'$}& The duration for which the UAV maintains  \\
\textbf{ }& maximum acceleration during the first stage of flight \\
\hline
\textbf{$r'(t)$}& The radius of obstacles in the UAV's map at time $t$.\\
\hline
\textbf{$y(t),v_y(t),$}& The position, velocity, and acceleration \\
\textbf{ $a_y(t)$ }& in the y-direction at time $t$.\\
\hline
\text{others}& Please refer to \figref{fig:scene} for more details \\
\hline
\end{tabular}

\label{tab1}
\end{center}
\end{table}

\subsection{Model derivation}

Under our assumptions, the UAV is a real-time system that makes decisions based only on current observations. The planner always hopes to produce a smoother curved flight, so we can divide the process of the UAV avoiding obstacles into two stages: obstacle avoidance and rapid direction return.

At the same time, since we assume that the UAV has a fixed target direction speed, this means that $v_x$ is always equal to $v_{safe}$, and the total time for the UAV to avoid obstacles is 

\begin{equation}
T = \frac{S-d}{v_{x}} 
\end{equation}

\textit{1) Obstacle avoidance: }
At this stage, the UAV needs to avoid obstacles directly in front. As per the previous assumptions, it is evident that the UAV, at the moment it perceives the obstacle in configuration space, is at a distance of $S-d$ from the obstacle. The subtraction of s accounts for the typical scenario where the UAV's camera is not positioned at the edge of the propellers. At the moment the obstacle is perceived, the UAV's collision boundary has already been pushed forward by a distance of $d$. Thus, in configuration space, the obstacle has expanded in the opposite direction by $d$.

Considering that the delay of the algorithm is $\tau$, in the scenario where the UAV is just about to avoid collision with the obstacle, the UAV should maintain a maximum vertical acceleration, $a_{max}$. With this in mind, we can obtain the following relationship:

\begin{equation}
\frac{1}{2}a_{max}(\frac{S-d}{v_{x,max}}-\tau)^2 = r+d 
\end{equation}

Moving $v_{x,max}$ to the left we can get,

\begin{equation} \label{eq:stage1}
v_{x,max} = \frac{S-d}{\tau - \sqrt{2\frac{r+d}{a_{max}}}} 
\end{equation}

This is the maximum safe speed that the UAV can accept without colliding with obstacles in the first stage, similar to the conclusion in [13].

\begin{figure}[htbp]
\centerline{\includegraphics[width=0.5\textwidth]{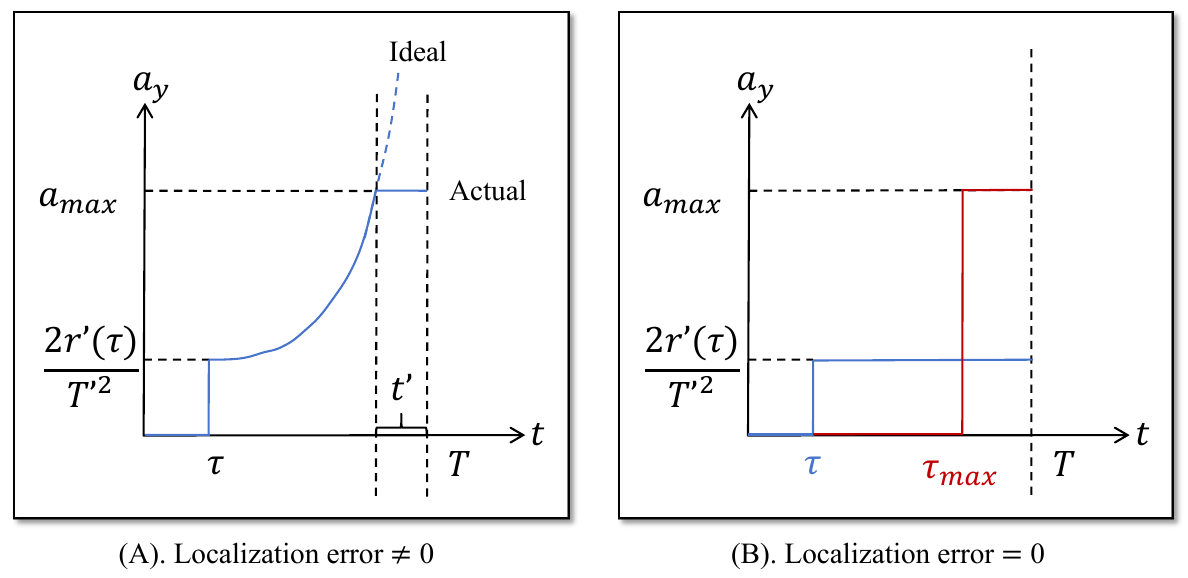}}
\caption{(A) The physical performance of the UAV constrains its acceleration to reach its maximum value after a certain period of time. (B) When the localization error is zero, the model degrades to an ideal state, and the acceleration remains constant. }
\label{fig:acc}
\end{figure}

To calculate the impact of the collision on the UAV's velocity in the second stage, we also need to determine the UAV's velocity in the y-direction at the moment of separation from the obstacle. As we assume that the UAV always deviates from the obstacle with in a smooth manner, the planner always generates trajectories that separate from the obstacles exactly, given a lateral velocity $v_x$, at any given time $t$ during the first stage, we assume the UAV's position to be $(x, y)$. Hence, we can obtain the following differential equation with respect to the UAV at that moment:

\begin{gather}
x = v_x t \\
r'(t) = r+d+e v_x t 
\end{gather}

if $t<\tau$:
\begin{equation}
    a_y = 0
\end{equation}
else if $t>\tau$:
\begin{equation}
\frac{1}{2}a_y(t) (T-t)^2 + v_y(t) (T-t) = r+d+ev_xt - y(t)
\end{equation}

Solving the above equation yields that, when $t>\tau$, the functions expressing the variation of $y$ and $v_y$ with respect to time can be represented as $Y(t)$ and $V_y(t)$:


\begin{equation}
	\begin{split}
	y(t) =& r'(T)-2ev_x(T'-t)ln(\frac{T'}{T'-t}) \\
	&- ev_x\frac{(T'-t)^2}{T'} -r'(\tau)\frac{(T'^2-t^2)}{T'^2}
	\end{split}
\end{equation}

\begin{equation}
v_y(t) =2 e v_x ln(\frac{T'}{T'-t}) - 2ev_x\frac{t}{T'} + \frac{2r'(\tau)t}{T'^2} 
\end{equation}

We can observe that as $t$ approaches $T$, $y$ tends towards $r + ev_xT$, which aligns with our expectations for the smooth flight of the UAV.

However, if we carefully observe the changes in $v$ and $a$, we will find that as $t$ approaches $T$, both $v$ and $a$ tend to infinity. 
This phenomenon is actually understandable. Due to the influence of localization error, even if the UAV is very close to the obstacle, the obstacle will still expand at the same speed. When the distance is shorter, the required acceleration will naturally be greater. However, in actual situations, the acceleration of the UAV is limited, so the acceleration will not continue to increase after it reaches the limit. The acceleration performance of the UAV is shown in \figref{fig:acc}, $t'$ denotes the duration of the UAV flying at maximum acceleration.

\begin{figure*}[htbp]
   \begin{overpic}[width=\textwidth]{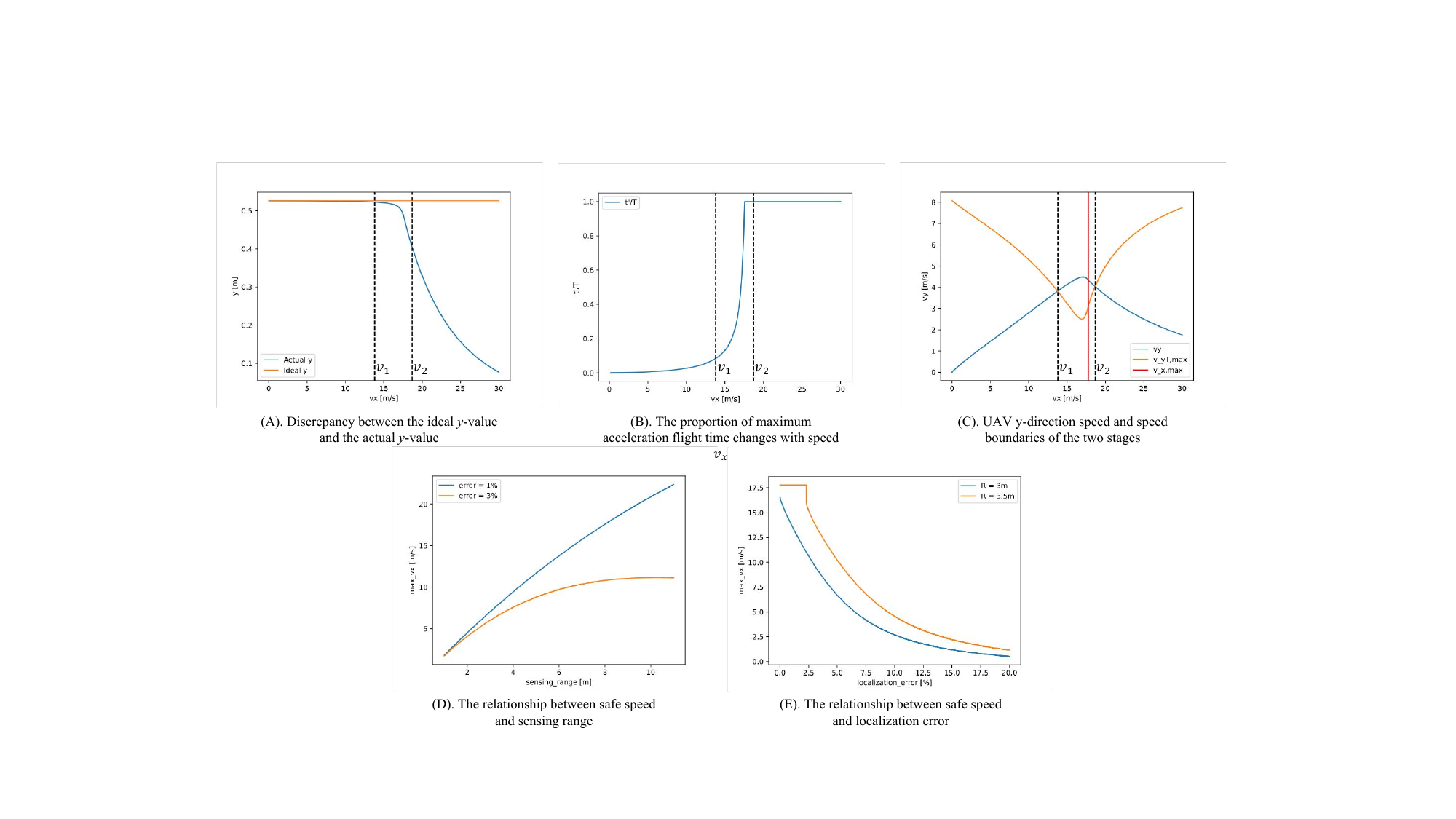} \small
   \end{overpic}
   \caption{(A)-(B) When $v < v_1$, the proportion of t' during the entire acceleration phase is very small, and the UAV's displacement in the y direction has a minimal deviation from the ideal. (C) The safe speed of the UAV is simultaneously constrained by two stages, leading to it being less than $v_1$. (D) Localization error causes the influence of sensing range on safe speed to deviate from linearity. (E) In high-density scenarios, localization error has a more pronounced impact on safe speed.
   }\label{fig:y_error}
\end{figure*}

The change node can be calculated based on the change in acceleration, and then the final speed and position can be calculated.
\begin{equation}
t' = \frac{2 e v_x}{a_{max} + 2 e v_x/T' }
\end{equation}
\begin{equation}
y(T) =y(T-t') + v_y(T-t') t' + \frac{1}{2}a_{max} t'^2
\end{equation}
\begin{equation}
v_{y}(T) = v_y(T-t') + a_{max} t'
\end{equation}

Although the acceleration is limited, it can be shown that $t'$ is usually a short time. This means that the resulting position deviation is almost negligible in most cases, as we will prove in the next section.

\textit{2) Rapid direction return: }
At this stage, the UAV needs to reduce its y-direction velocity to zero within a finite distance $L$. Based on the calculated $v_y(T)$ and $y(T)$ during the first stage, along with the restriction on jerk, we can easily obtain the following expression:

\begin{equation}
y_{max} < L
\end{equation}

if $t<2a_{max} / j_{max}$:

\begin{equation}
a_y = a_{max} - j_{max} t
\end{equation}

else:

\begin{equation}
a_y = -a_{max}
\end{equation}

We can get:

\begin{equation} \label{eq:stage2}
y(T) + \frac{2a_{max}v_{y}(T)}{j_{max}} + \frac{2a^3_{max}}{j^2_{max}} + \frac{v_{y}^2(T)}{2a_{max}} < L  
\end{equation}

When Equation \eqref{eq:stage2} is equal, this constraint determines the maximum safe speed to avoid colliding with obstacles in the second stage, where $v_y(T)=v_{y,max}(T)$.
The values of $v_{y,max}(T)$ and $v_{x,max}$ jointly constrain the maximum safe speed of the UAV. In the solving process, we utilize the conditions $v_x < v_{x,max}$ and $v_{y}(T) < v_{y,max}(T)$ for numerical computation.

\section{Model analysis}

In this section, we will analyze the model based on the assumption of the UAV performing a jungle traversal mission. All analysis results are based on the typical values of the parameters, as shown \tabref{tab2}. Subsequent analyses will involve single-variable modifications based on this parameter set.

\begin{table}[htbp]
\caption{Default values of flight parameters.}
\begin{center}
\resizebox{0.5\textwidth}{11pt}{
    \begin{tabular}{|c|c|c|c|c|c|c|c|c|}
    \hline
    \textbf{ parameter } & \textbf{ $r$ }& \textbf{ $d$ }& \textbf{ $a_{max}$ }& \textbf{ $j_{max}$ }& \textbf{ $R$ }& \textbf{ $e$ }& \textbf{ $S$ }& \textbf{ $\tau$ }  \\
    \hline 
    \textbf{ value } & $0.1m$ &$0.37m$ &$20m/s^2$&$120m/s^3$&$3m$&$0.01$&$6m$&$0.01s$\\
    \hline
    \end{tabular}
}
\label{tab2}
\end{center}
\end{table}

The typical values for localization accuracy, perception range, and their corresponding delays were selected based on [20], [22]. The safety distance was measured using the simulator.

\subsection{Analysis of Flight State Variations}

The first question is how the limitations on the first-stage velocity components, $v_x$ and $v_y$, obtained in the previous section, are reflected in the maximum safe speed. It is easy to understand that $v_x < v_{x,max}$ ensures that the UAV does not collide with obstacles in the first stage as long as the flight speed does not exceed $v_{x,max}$. This is represented by the left side of the red line in \figref{fig:y_error}.(C). 

Regarding $v_v(T) < v_{y,max}(T)$, by plotting the curves of $v_v(T)$ and $v_{y,max}(T)$ as a function of $v_x$ in the graph, we can find two intersection points, $v_1$ and $v_2$. This restricts the UAV speed to be less than $v_1$ or greater than $v_2$. Taking into account the limitation of avoiding obstacles in the first stage, if $v_{y,max}(T) > v_2$, the UAV speed is limited to the range of $0$ to $v_1$ and $v_2$ to $v_{y,max}(T)$. 

However, in reality, even if this situation occurs, when the velocity $v_x$ exceeds $v_2$, the UAV has a significant amount of time in which it cannot reach the desired acceleration, resulting in a large discrepancy between its actual position in the y-direction and the ideal position, as shown in \figref{fig:y_error}.(A)-(B). Therefore, in planning, it is unreasonable for the UAV model to overlap with the inflated obstacles for a significant amount of time, and thus the speed is restricted to the range of 0 to $v_1$. In other words, under the limitation of the upper obstacle, the maximum safe speed for the UAV is $v_1$.

As for why $v_y(T)$ exhibits an increase followed by a decrease with respect to $v_x$, we can divide the first stage of the flight into two parts. The first part follows the theoretical results and controls the acceleration. The acceleration curve in this part aligns with the theoretical calculations and takes a time of $T-t'$. The second part occurs when the required acceleration reaches the UAV's limit, and the UAV can only fly with its best effort. In this part, the UAV maintains the maximum acceleration value, $a_{max}$, for a duration of $t'$. The total time for both parts is $T = \frac{S-d}{v_{x}}$.

We can imagine that as $v_x$ increases, $t'$ should first increase and then decrease. When the UAV's speed is low, as the speed increases, the required acceleration for the UAV to avoid obstacles becomes higher. This results in an increasing proportion of $t'$ within the total time $T$, and consequently, $t'$ increases. However, as the speed further increases, $t'$ gradually occupies the entire $T$. At the same time, the UAV becomes less capable of avoiding obstacles effectively. A faster speed means a shorter avoidance time $T$ itself, which leads to a decrease in $t'$. This actually causes the decay of $v_y(T)$ as well. By plotting the ratio of $t'$ to $T$ as a function of $v_x$, as shown in Figure 1, we can observe that this transition indeed occurs between $v_1$ and $v_2$. 

\textbf{This means that the region between $v_1$ and $v_2$ is actually a transitional period where the UAV gradually deviates from ideal control, and its performance approaches its limits.}

\begin{figure*}[htbp]
   \begin{overpic}[width=\textwidth]{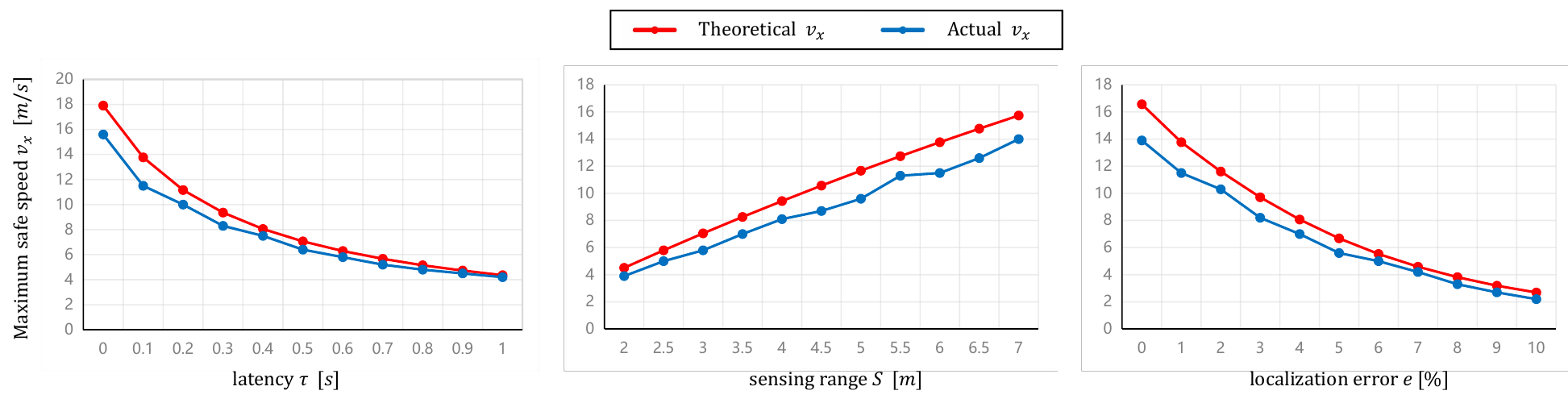} \small
   \end{overpic}
   \caption{Comparison between theory and actual measurement results. In the case of the highest flight speed of 18m/s, our model maintains an error below $20\%$, and the lower the speed, the more accurate the model becomes.
   }\label{fig:result}
\end{figure*}

Another question is whether the deviations in the UAV's flight, since it is always non-ideal, would result in direct collisions with obstacles. In fact, this deviation is minimal when $v < v_1$. If we plot the ideal and actual positions on a graph, with the lateral velocity $v_x$ as the x-axis, we obtain \figref{fig:y_error}.(A). It can be observed that the difference between the theoretical and actual values is very small. This difference essentially arises from the continuous expansion of obstacles, leading to misjudgments by the UAV. The ``collision" portion that occurs in the mapping process actually involves only a short overlap in time, rather than actual physical collisions. In cases where the safety distance is large, there may not even be any overlap.


\begin{figure}[htbp]
\centerline{\includegraphics[width=0.5\textwidth]{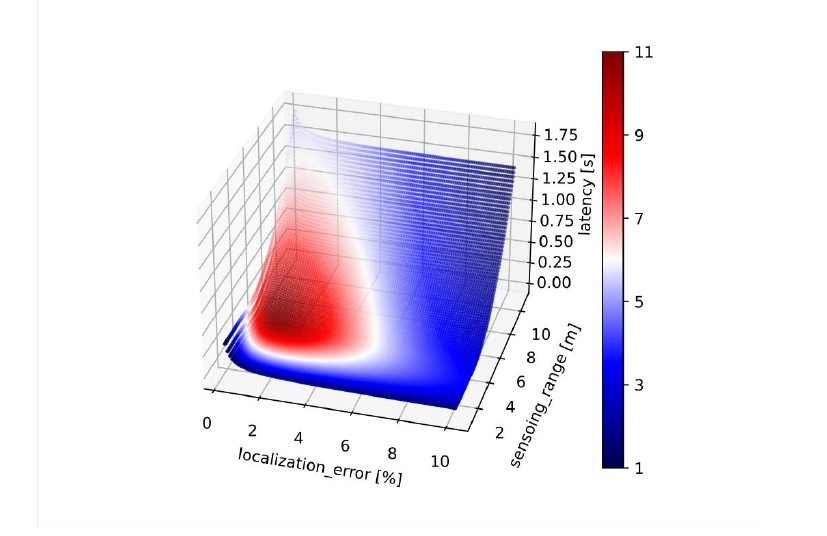}}
\caption{The coupling effect of multiple parameters causes the peak value of maximum safe speed to not occur at the position of minimum localization error or maximum sensing range.}
\label{fig:couple}
\end{figure}

\subsection{Analysis of UAV Parameter Effects}

Regarding the sensing range, according to \figref{fig:y_error}.(D), when the localizaiton error is small, the UAV's speed increases almost linearly with an increase in the sensing range. This aligns with the representation of the limit in the first stage where $v_x < v_{x,max}$. 

However, as the localizaiton error increases, the improvement in flight speed gradually deviates from linearity with an increase in the sensing range. This is because a larger sensing range causes the UAV to observe obstacles earlier, which also means that the expansion of obstacles occurs earlier. As a result, the UAV's y-direction speed becomes excessive. The boundaries of the second stage gradually restrict the growth of safe speed.

When it comes to the localization error, based on \figref{fig:y_error}.(E), its impact on the maximum safe speed is evident. However, in most cases, the values of the positioning error are small, which is reflected in the first half of the curve in the graph. Furthermore, if the obstacle density is low, such as in the case of $R = 3.5m$, we can observe that at small error values, the maximum safe speed is not affected by the localization error. This suggests that the boundaries of the second stage have not yet come into effect. 

In an extreme case where the environment is infinitely open with a negligible obstacle density ($R$ tends to infinity) and only a single pillar is present, there exists only the first stage boundary of Equation \eqref{eq:stage1}. In this case, the model degenerates into the basic model, independent of localizaiton accuracy. However, as the distance between obstacles decreases, the influence of localization error on the safe speed gradually becomes significant.

In practical scenarios, the effects of localization error, sensing range, and computational delay are interrelated. Improvements in localization error and sensing range result in greater computational delays. Due to different technological approaches, the relationship between delay and changes in localization error and sensing range varies. Based on the equation $\tau = \tau_{depth} + \tau_{localization}$, we depict a potential surface in \figref{fig:couple}, illustrating the overall delay as a function of localization error and sensing range. The x-axis represents localization error, the y-axis represents sensing range, and the z-axis represents the computational delay, with the surface's parameters being rough estimates. Overall, it demonstrates that reducing localization error and expanding sensing ranges lead to increased delays. The color of points on the surface indicates the magnitude of the UAV's maximum safe speed. From the figure, we can observe that the peak of the maximum safe speed does not always occur at the location with minimal localization error and maximal sensing range. Pursuing excessively high values for a single parameter may result in a loss of overall performance. 

By utilizing our model, UAV designers can select the most suitable combination of parameters across different software and hardware to achieve maximized safe flight speed.

\section{EXPERIMENTS}

\subsection{Experimental setup}

\begin{figure}[htbp]
\centerline{\includegraphics[width=0.5\textwidth]{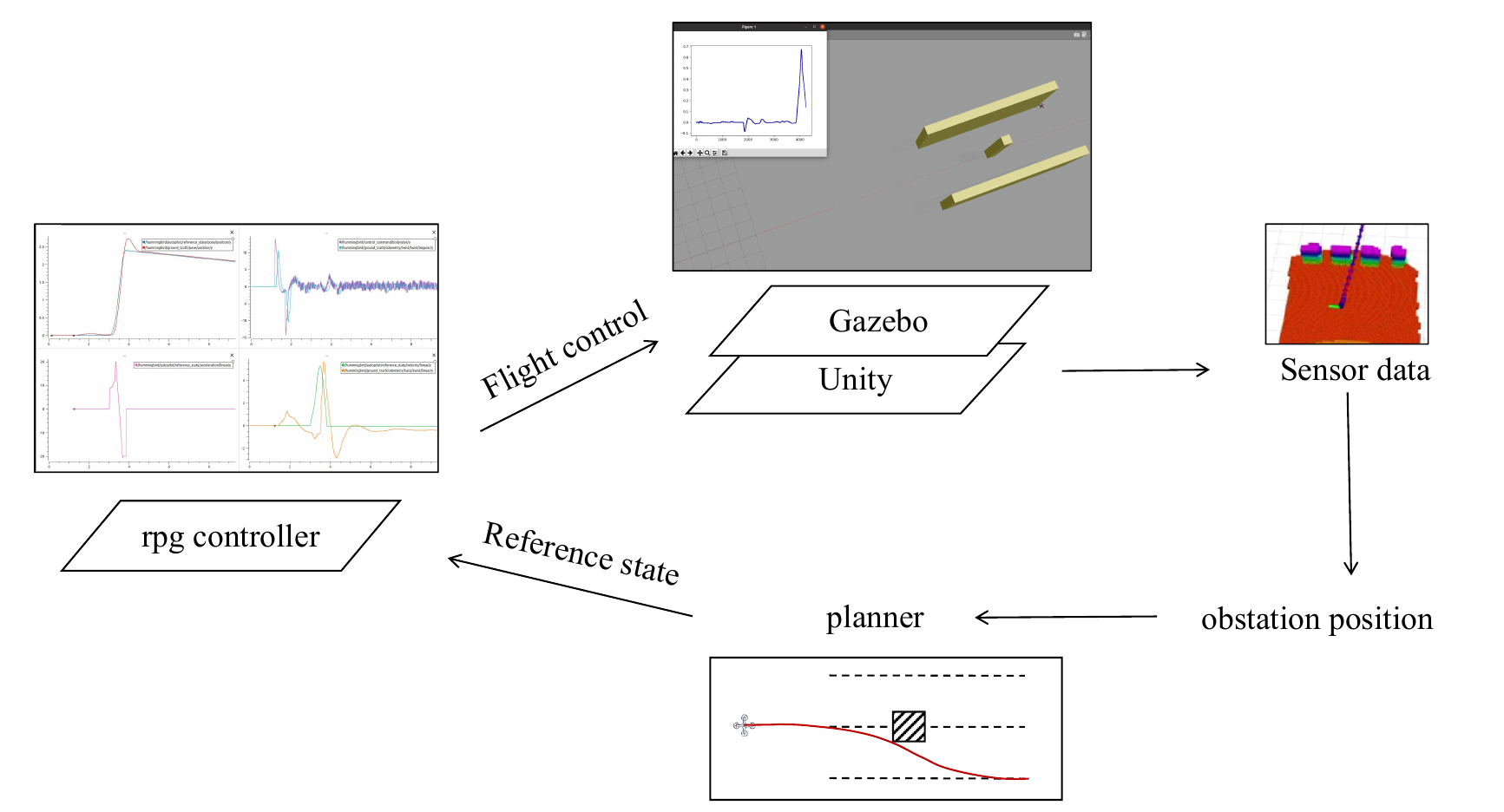}}
\caption{Building upon the flightmare simulation [23], we abstracted the observation data of the UAV and directly fed the obstacle states into the planner, resulting in the generation of flight trajectories that align with our expectations.}
\label{fig:exp}
\end{figure}

In the experiment, Flightmare is employed as the simulation environment [23], and the overall structure of the environment is illustrated in \figref{fig:exp}.

During the experiment, to achieve controllable latency, we abstracted the observation and planner components of the UAV. The UAV perception module directly outputs the observed state of obstacles (length and distance) instead of depth images. The planner component calculates the acceleration based on the obstacle information and current position, and further converts it into a trajectory provided to the controller. The controller utilizes the control algorithm proposed in [24], [25].


The scene design is illustrated in \figref{fig:exp}. The UAV takes off from the coordinate origin, and the obstacle is positioned 25m away from the UAV. A reserved distance is provided to allow the UAV to accelerate and ensure that it reaches the predefined velocity ($v_x$) before encountering the obstacle. This setup ensures that the UAV can stably fly while maintaining the desired velocity after detecting the obstacle.

To test the maximum safe speed of the UAV, we need to scan the $v_x$ parameter under different settings until we reach the critical point where the UAV can just pass through the obstacle. Determining whether the UAV has collided is actually straightforward. We design the simulation environment to include three obstacles. This way, if a collision occurs between the UAV and any obstacle, there will be a noticeable change in the velocity along the x or y direction. Therefore, our design for the UAV's safety distance is precise, matching the size of the UAV exactly.

The main reason for simplifying the UAV sensors is to reduce the delay in the perception and planning phase of the UAV, allowing us to have a greater range of manual control over the calculation delay. We approximate that, without adding any additional delay, the system delay is considered to be zero.

\subsection{Experimental result}

After conducting tests, we plotted the relationship between the theoretical maximum safe speed and the actual tested maximum safe speed for different parameters in \figref{fig:result}. The default parameters are set as indicated in \tabref{tab2}. In the graph, from left to right, it represents the variation of the maximum safe speed with respect to the delay, localization error, and sensing range. Upon observation, as the flight speed increases, the discrepancy between the predicted speed by the model and the actual speed gradually increases. Under the highest flight speed of $18m/s$, the difference between our prediction and the actual result is less than $20\%$.

The main reason for the error is that during the second stage of the UAV flight, it is necessary to rapidly transition from maximum forward acceleration to maximum reverse acceleration. This rapid change causes a certain amount of overshoot in the UAV control process, resulting in the actual yaw distance being slightly larger than the theoretical distance.


\section{Conclusions}

In this work, we investigated the impact of parameters such as localization error, sensing range, and computational latency on the maximum safe speed of UAVs. By assuming avoidance and correction mechanisms for obstacles in typical scenarios, we demonstrated how parameters like localization error affect UAV flight. The results showed that as obstacle density increases and the flight space for UAVs decreases, the influence of localization error on the maximum safe speed becomes increasingly significant. Through a coupled analysis of multiple parameters, we discovered that optimizing a single metric does not always guarantee the optimal overall flight speed of the system. Joint calculations considering multiple parameters can provide better guidance for UAV design. We validated our conclusions in a simulator, where the model predicted the safe speed within an error margin of $20\%$ when the maximum flight speed reached $18 m/s$. In the future, we will further verify the flight performance of UAVs on real unmanned aircraft.



\end{document}